\documentclass[sigconf, authorversion, screen]{acmart}
\AtBeginDocument{%
  \providecommand\BibTeX{{%
    \normalfont B\kern-0.5em{\scshape i\kern-0.25em b}\kern-0.8em\TeX}}}

\copyrightyear{2021}
\acmYear{2021}
\setcopyright{acmcopyright}\acmConference[FDG'21]{The 16th International Conference on the Foundations of Digital Games (FDG) 2021}{August 3--6, 2021}{Montreal, QC, Canada}
\acmBooktitle{The 16th International Conference on the Foundations of Digital Games (FDG) 2021 (FDG'21), August 3--6, 2021, Montreal, QC, Canada}
\acmPrice{15.00}
\acmDOI{10.1145/3472538.3472579}
\acmISBN{978-1-4503-8422-3/21/08}

\begin{document}

\title{Using Machine Learning to Predict Game Outcomes Based on Player-Champion Experience in League of Legends}

\author{Tiffany D. Do}
\affiliation{%
  \institution{University of Central Florida}
  \city{Orlando, Florida}
  \country{United States}}
\email{tiffanydo@knights.ucf.edu}

\author{Seong Ioi Wang}
\affiliation{%
  \institution{University of Texas at Dallas}
  \city{Richardson, Texas}
  \country{United States}}
\email{seongiwang@gmail.com}

\author{Dylan S. Yu}
\affiliation{%
  \institution{University of Texas at Dallas}
  \city{Richardson, Texas}
  \country{United States}}
\email{dylanyu461@gmail.com}

\author{Matthew G. McMillian}
\affiliation{%
  \institution{University of Texas at Dallas}
  \city{Richardson, Texas}
  \country{United States}}
\email{matthewgmcmillian@gmail.com}

\author{Ryan P. McMahan}
\affiliation{%
  \institution{University of Central Florida}
  \city{Orlando, Florida}
  \country{United States}}
\email{rpm@ucf.edu}
\renewcommand{\shortauthors}{Do, et al.}

\begin{abstract}
  League of Legends (LoL) is the most widely played multiplayer online battle arena (MOBA) game in the world. An important aspect of LoL is competitive ranked play, which utilizes a skill-based matchmaking system to form fair teams. However, players' skill levels vary widely depending on which champion, or hero, that they choose to play as. In this paper, we propose a method for predicting game outcomes in ranked LoL games based on players' experience with their selected champion. Using a deep neural network, we found that game outcomes can be predicted with 75.1\% accuracy after all players have selected champions, which occurs before gameplay begins. Our results have important implications for playing LoL and matchmaking. Firstly, individual champion skill plays a significant role in the outcome of a match, regardless of team composition. Secondly, even after the skill-based matchmaking, there is still a wide variance in team skill before gameplay begins. Finally, players should only play champions that they have mastered, if they want to win games.
\end{abstract}




\maketitle

\section{Introduction}
League of Legends (LoL), a popular computer game developed by Riot Games, is currently the most widely played Multiplayer Online Battle Arena (MOBA) \cite{Aung2018} game in the world. In 2019, there were eight million concurrent players daily \cite{leaguestats}, and the player base has continued to grow since its release in 2009. A core aspect of LoL is competitive ranked gameplay. In typical ranked gameplay, ten human players are matched together to form two teams of approximately equal skill. These two teams, consisting of five players each, battle against each other to destroy the opposing team's base. 

Fair matchmaking is crucial for player experience \cite{Veron2014}. In 2019, Riot Games stated that ranked matchmaking should be as fair as possible \cite{ranked_2019}. This goal has persisted throughout the history of the game. In 2020, Riot Games stated that some of their main goals for the year were to preserve competitive integrity \cite{lookback_2020} and improve matchmaking quality \cite{league_matchmaking} for ranked games. In order to create fair matches between players of approximately equivalent skill level, matchmaking is determined using an Elo rating system, similar to the one originally used by chess players \cite{Mora-Cantallops2018}. Although this matchmaking system has improved in recent years (c.f., \cite{ranked_2018, ranked_2019, league_matchmaking}), it does not consider players' champion selections when forming matches. LoL has over 150 playable characters, known as champions, that have their own unique playstyles and abilities \cite{Do2020}. Players select a champion at the start of every match after the matchmaking algorithm has formed teams. However, players will often perform better on some champions than on others due to their differing levels of mechanical expertise, which is defined as a player's knowledge of their champion's abilities and interactions \cite{Donaldson2017}. Higher levels of mechanical expertise on particular champions allow players to make quicker and better judgments, which are essential in the game's fast paced environment. Since mechanical expertise plays such a large impact on a player's own performance, it can therefore cause a similar impact on the match's outcome. 

In this paper, we introduce a machine learning model based on a deep neural network (DNN) that can predict ranked match outcomes based on players' experience on their selected champion (i.e., player-champion experience). We show that the outcome of a ranked match can be predicted with over 75\% accuracy after all players have selected their champions. This occurs before the match actually begins. Our results indicate that individual champion skill plays a significant role in the outcome of a match, regardless of team composition. Additionally, current matchmaking may not form teams of approximately equal skill level in game because matchmaking is done before players select their champions. Our results also imply that players should only play champions that they have mastered in order to win ranked games. This can decrease gameplay variety as players typically master only a few champions out of the 150+ champions available in the roster \cite{Chen2017, Do2020}.

\section{Background and Related Work}

\subsection{Ranked Gameplay in League of Legends}
Ranked gameplay in LoL is based on seasonal gameplay, similar to traditional sports. During each season, LoL separates players into six main divisions known as Leagues based on skill level. These Leagues are known as Iron, Bronze, Silver, Gold, Platinum, and Diamond \cite{Li2020}. There are a few distinctions past Diamond. However, we do not consider these players for the purposes of this player as they make up a minuscule portion of the playerbase. In order to rank up and enter higher Leagues, players must win ranked matches. In ranked matches, two teams of five players each are formed using a matchmaking algorithm. 

For ranked matches, LoL utilizes a matchmaking algorithm based on a hidden Elo rating system, similar to the one originally used by chess players \cite{Mora-Cantallops2018}. This matchmaking algorithm has been continuously improved upon since the release of the game and is used to create matches consisting of players of approximately equal skill level \cite{ranked_2018, ranked_2019, league_matchmaking}. In 2020, Riot released a new matchmaking algorithm that better identifies player skill level \cite{league_matchmaking}.

When a player decides to play a ranked match, they will be matched with nine other players (via the matchmaking algorithm) who are then separated into two teams. Ranked matches are played with a minimum of eight strangers. Players may choose to enter matchmaking alone or with a similarly ranked partner, in which case they will be placed on the same team as their partner \cite{Mora-Cantallops2018}.

Since mechanical expertise is important to a player's performance, it seems logical that players will select champions that they have high expertise on. However, this may not always be the case. An important factor of champion selection is team composition. An optimal team has a combination of different roles in the game \cite{Kou2014}. Certain champions have different roles, with some examples being "Tank" (champions with high defense), "Mage" (long ranged champions with burst damage and crowd control), and "Support"(champions with support ability) \cite{Gao2017}. However, since players are playing with strangers, they cannot predetermine optimal team compositions. For instance, if no players on the team typically play "Tank" champions, one player may be forced to play a "Tank" champion, even though they may not have high mechanical expertise on it.

\subsection{LoL Game Outcome Predictors}
In this section, we describe previous work on LoL game outcome predictors. Most MOBA game outcome predictors are focused on two popular MOBAs: LoL and Dota 2. For the purposes of this paper, we focus only on LoL predictors due to the differences in gameplay between LoL and Dota 2. For instance, Chen et al. \cite{Chen2017} indicated that LoL has more diverse skill compositions than Dota 2. 

There are several predictors that use pre-match data (i.e., before the match begins) like our predictor. However, the majority of these pre-match predictors have an accuracy around 55-60\% (e.g., \cite{Chen2017, Lin2016, Kim2019}). Similar to our work, Chen et al. \cite{Chen2017} used player-champion experience to predict game outcomes. Using a combination of player-champion experience, champion experience, and player experience, they predicted game outcomes with an accuracy of 60.24\% using logistic regression. They found that player-champion experience was more influential than both champion and player experience on game outcomes. We further expand on this work by investigating multiple features of player-champion experience, since Chen et al. \cite{Chen2017} used only one feature to describe player-champion experience.

Within-match predictors predict the winner while the match is ongoing. Typically, as the game time elapses, the accuracy increases. Data such as kill difference, gold difference, and towers destroyed can help determine which team has the lead and will ultimately win the game. Lee et al. \cite{Lee2020} found that within-match data can predict the winner with 62.26\% accuracy at 5 minutes elapsed and 73\% accuracy at 15 minutes elapsed using Random Forest (RF). Similarly, Silva et al. \cite{Silva2018} predicted the winner with an accuracy of 63.91\% at 5 minutes elapsed and 83.54\% at 25 minutes elapsed using a recursive neural network. Unlike these predictors, our model does not rely on within-match data and instead predicts the match outcome before the match has begun, deviating entirely in functionality and application.

Ani et al. \cite{Ani2019} found that pre-match data can predict the game outcome of professional LoL games with very high accuracy (>90\%) using RF trees and ban data (each team is allowed to ban up to five champions). However, we focus on ranked gameplay rather than organized professional gameplay. Unlike professional games, ranked games are designed to have fair teams consisting of strangers and support all varieties of skill levels. Additionally, professional games contain a much different playerbase from ranked games, as they are only played among a couple dozen players.

\section{Methodology}
\subsection{Dataset}
Riot Games provides a public API \cite{RiotAPI} containing several types of data pertaining to players and their matches. This data contains features such as champion mastery, games played per season, and win rates. Since our methodology involves training machine learning models with data from ranked matches, we used this API to query random players and filter their most recent ranked match. Using this method, we pulled a total of 5000 unique ranked matches for our dataset. Any duplicate matches were removed. From these matches, we recorded general match information, as well as information about each player, specifically because they describe a player's experience on a given champion:

\begin{itemize}
    \item \textbf{Champion mastery points} - An integer $\in [0, \infty]$ approximating the player's lifetime experience on the given champion \cite{Do2020}. Champion mastery points are accrued by playing matches on the champion. The mean champion mastery points of players in our dataset was 122,368.44, while the max value was 9,364,624. 
    \item \textbf{Player-champion win rate} - A percentage (0.0 to 1.0) indicating the player's victory ratio for ranked games played on the given champion during the 2020 ranked season. 
    \item \textbf{Season total number of games played on the champion} - An integer $\in [0, \infty]$ signifying the total number of ranked games the player has played on the given champion during the 2020 ranked season. The mean season total number of games of players in our dataset was 58.95, while the max value was 1,819. 
    \item \textbf{Number of recent games played on the champion} - An integer $\in [0, 20]$ representing the number of ranked games that the player has played on the champion within their last 20 ranked games during the 2020 ranked season. 
\end{itemize}

These features represented each player's experience levels on their chosen champion and helped us determine whether player-champion experience can predict match outcomes. Overall, our methodology allowed us a large pool of data to build upon and use for our model, while also stripping away personally identifiable data from each player.

\subsection{Feature Selection}
Using Pearson's correlation test, we discovered that player-champion win rate and champion mastery points were the only features that correlated with the outcome of a match, and that the total number of games played and number of recent games played on a particular champion had no effect on the outcome of the match. Furthermore, during preliminary analysis, our models performed worse when the other features were included in the training dataset.

For each match, we derived additional features to better reflect each team's overall player-champion experience. For both teams, we added the team's average, median, coefficient of excess kurtosis \cite{zwillinger1999crc} (i.e., how normal the distribution is), coefficient of skewness \cite{zwillinger1999crc}, standard deviation, and  variance of the players' champion win rate and of their champion mastery points. The final training dataset contained 44 features per match (2 features per player for 10 players and 12 features per team).

\subsection{Models}
We chose to explore several machine learning models based on the success of previous work that also investigated MOBA game outcomes \cite{Anshori2018, Semenov2017, Ani2019, Lin2016, Lee2020}. We investigated Support Vector Classifiers (SVC) \cite{Kecman2005}, k-Nearest Neighbors (kNN) \cite{ISL}, Random Forest (RF) trees \cite{ESLII}, Gradient Boosting (GBOOST), and Deep Neural Networks (DNN) \cite{ESLII}. In this section, we briefly explain each model, as well as any parameters that we used for our implementations.

\subsubsection{Support Vector Classifier}
An SVC is a form of Maximal Margin Classifier that attempts to form a $(p-1)$ dimensional separable hyperplane on a $p$ dimensional feature space \cite{ISL}. SVCs intentionally misclassify training observations in an attempt to improve the classification of remaining observations \cite{ISL}. We used scikit-learn's \cite{scikit-learn} implementation of SVC with the following modified parameters: \texttt{C=8}, \texttt{coef0=1}, \texttt{kernel='poly'}, and \texttt{tol=1e-2}. We opted to use a polynomial kernel over a radial-basis-function kernel \cite{ISL} because the polynomial kernel showed better classification accuracy.

\subsubsection{k-Nearest Neighbors}
The kNN \cite{DAPar} algorithm takes a given record $x$ and compares it with the $k$ closest records from its corresponding dataset, with closeness defined using some distance measure. We used scikit-learn's \cite{scikit-learn} implementation of kNN with the following modified parameters: \texttt{leaf\_size=5}, \texttt{n\_neighbors=600}, \texttt{p=1} and \texttt{weights='distance'}. For this model in particular, we noticed that including more features caused our overall prediction accuracy to drop dramatically. In order to circumvent the curse of dimensionality \cite{ISL}, we reduced our feature space to only include the team's average win rate.

\subsubsection{Random Forest Trees}
RF trees \cite{ESLII} use bagging to create multiple base decision tree classifiers and aggregate them into a single model. This method can yield a model with less variance and overfitting in comparison to classic decision trees. We used scikit-learn's \cite{scikit-learn} implementation of RF trees \cite{ESLII} with the following modified parameters: \texttt{n\_estimators=350}, \texttt{min\_samples\_leaf=3}, \texttt{min\_samples\_split=10}, and \texttt{max\_features=7}.

\subsubsection{Gradient Boosting}
GBOOST is another ensemble method that sequentially adds predictors to its ensemble and tries to fit each new predictor based on the residual errors of the previous predictor \cite{geron2017hands-on}. We used scikit-learn's \cite{scikit-learn} implementation of GBOOST, whose default estimator is a Decison Tree  \cite{geron2017hands-on}, with the following modified parameters: \texttt{learning\_rate=0.14} and \texttt{n\_estimators=55}.

\subsubsection{Deep Neural Networks}
We utilized Keras \cite{chollet2015keras}, a Python deep learning API, to build our model. We decided to go with a pyramid network architecture based on examples described in \cite{geron2017hands-on}, as we noticed that it performed better than networks with hidden layers with the same number of neurons in our preliminary experiments.
We used the following network topology to define our model:
\begin{itemize}
    \item A flattened input layer resulting in a $1 \times 44$ output.
    \item Alternating dropout, normalization, and dense layers for a total of 15 layers (5 dropout, 5 normalization, and 5 dense layers). Each group of alternating layers had 160, 128, 64, 32, and 16 neurons, in that order.
    \begin{itemize}
        \item Each dropout layer had a dropout rate of $0.69\%$.
        \item Each normalization layer utilized batch normalization.
        \item Each dense layer used Exponential Linear Unit (ELU) activation \cite{geron2017hands-on}, He initialization \cite{geron2017hands-on}.
    \end{itemize}
    \item A $1\times 1$ dense layer with Sigmoid activation
\end{itemize}
Notably, we used ELU activation, He initialization and batch normalization to deal with unstable gradients \cite{geron2017hands-on}, which led to more accurate model predictions. We also added dropout layers to prevent overfitting \cite{srivastava2014dropout}.

\section{Experiments and Results}

\subsection{Experimental Settings}
For our experiments, we split our dataset into a training and testing dataset. For the training dataset, 80\% of the matches (4000 matches) were randomly chosen, while the other 20\% (1000 matches) were chosen for the testing dataset. We utilized stratified k-fold validation \cite{CANO200790} with $k=10$ to evaluate our models. Stratified k-fold validation tries to keep the class distribution of the testing dataset as similar to the actual class distribution as possible, which can lead to better accuracy. Each model's average accuracy, its standard deviation, and its standard error are shown in Table \ref{table:model-results}.

\begin{table}[]
\caption{The performance of each machine learning model, including the accuracy, 95\% confidence interval (CI), standard deviation, and standard error.}
\begin{tabular}{lcccc}
\toprule
Model         & Mean Accuracy $\pm$ CI        & Std. Dev. 		 & Std. Error 	\\ \hline
SVC           & 74.3\% $\pm$ 1.21\%           & 1.7\%       	   & 0.54\%  	\\
KNN           & 72.7\% $\pm$ 1.23\%           & 1.2\%              & 0.38\%  	\\
RF            & 74.7\% $\pm$ 1.2\%	          & 2.0\%              & 0.63\%  	\\
GBOOST        & 75.4\% $\pm$ 1.19\%           & 5.25\%             & 1.66\%  	\\
DNN           & 75.1\% $\pm$ 1.2\%            & 1.9\%              & 0.60\%  	\\
\bottomrule
\end{tabular}
\label{table:model-results}
\end{table}

\begin{figure}  
    \centering
    \includegraphics[width=3.5in]{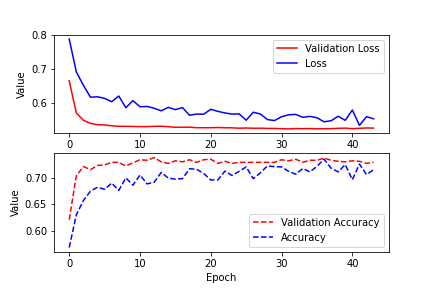}
    \caption{The loss and accuracy of our deep neural network while training and validating.}
    \label{fig:DDModule}
\end{figure}

\subsection{Results}
From our experiment results (see Table \ref{table:model-results}), we can see that all models predicted the winner with relatively high accuracy (>70\%). GBOOST had the highest accuracy with 75.4\%, which was marginally higher than our DNN model's of 75.1\%. However, the standard error of GBOOST was much greater than all of our other models. Therefore, we consider our DNN model to be most suitable for game outcome prediction since its standard error was low. Interestingly, from Figure \ref{fig:DDModule}, we can see that the validation accuracy is higher than the training accuracy of our DNN model. This is a byproduct of using dropout layers \cite{geron2017hands-on}. Using dropout layers artificially increases the difficulty of learning for the network during training, with the intent to increase test/validation accuracy. 

\section{Limitations and Future Work}
It is important to note that our work is limited to our specific domain. We only analyzed ranked matches from the North American (NA) server that occurred in 2020 within ranks Iron to Diamond. These results may not be similar for other servers such as the Oceania (OCE) server, due to differences between the playerbases, and may not reflect highly skilled play (top 1\%). Additionally, the playerbase may change over time. In the future, it would be interesting to investigate other regional servers (e.g., EUNE, EUW, OCE) to determine if player demographics have an effect on prediction accuracy. Furthermore, it may be of value to determine player-role experience and determine if this could serve as a similar predictor of a player's skill.

\section{Conclusion}
In this paper, we introduced a DNN machine learning model that can predict the accuracy of ranked LoL matches with an accuracy of 75.1\% using player-champion experience. To reflect each player's champion experience in a ranked match, we extracted multiple features regarding player statistics on their chosen champion. However, we found that only champion mastery points and ranked win rate on the selected champion were suitable for predicting game outcomes. We used this data to create summary features that represented the overall player-champion experience of the team. Using these summary features, we also found high accuracy (73\%-75\%) using other machine learning models, such as random forest trees and support vector classifiers, showing that player-champion experience can be a strong indicator of team performance in general.

Our results offer insight into the importance of champion selection and matchmaking design. Since we are able to determine the winner with reasonable accuracy after champion selection, this implies that even after the skill-based matchmaking, there is still a wide variance in team skill before gameplay begins. Fair matchmaking is important for player experience \cite{Veron2014}, and ranked matches in League of Legends are designed to be as fair as possible \cite{ranked_2019}. However, the current skill-based matchmaking system does not consider a player's champion selection, which can widely vary their skill level and thus cause the match to feel imbalanced. We show that the outcome of a ranked match can be determined with relatively high accuracy before gameplay begins based solely on player-champion experience. Although this problem cannot be resolved with the current matchmaking system, our results imply that players should limit their choices to champions that they have already mastered, regardless of team composition. However, most players can only master a handful of champions \cite{Chen2017}. This can be an issue for gameplay variety as players may choose not to experience the large roster of champions, in favor of playing the same few champions in order to win games.

% \begin{acks}
% \end{acks}

\bibliographystyle{ACM-Reference-Format}
\bibliography{sample-base}


\end{document}